\documentclass[conference]{IEEEtran}
\IEEEoverridecommandlockouts
\usepackage{cite}
\usepackage{amsmath,amssymb,amsfonts}
\usepackage{algorithmic}
\usepackage{graphicx}
\usepackage{textcomp}
\usepackage{xcolor}
\usepackage{makecell}
\usepackage{subfigure}
\def\BibTeX{{\rm B\kern-.05em{\sc i\kern-.025em b}\kern-.08em
    T\kern-.1667em\lower.7ex\hbox{E}\kern-.125emX}}
\begin{document}

\title{HDRSDR-VQA: A Subjective Video Quality Dataset for HDR and SDR Comparative Evaluation\\
}

\author{Bowen Chen, Cheng-han Lee$^{*}$, Yixu Chen$^{*}$, Zaixi Shang, Hai Wei, and Alan C. Bovik,~\IEEEmembership{Life Fellow,~IEEE}%
\thanks{$^{*}$ Cheng-han Lee and Yixu Chen contributed equally to this work.}%
}

\maketitle

\begin{abstract}
We introduce HDRSDR-VQA, a large-scale video quality assessment dataset designed to facilitate comparative analysis between High Dynamic Range (HDR) and Standard Dynamic Range (SDR) content under realistic viewing conditions. The dataset comprises 960 videos generated from 54 diverse source sequences, each presented in both HDR and SDR formats across nine distortion levels. To obtain reliable perceptual quality scores, we conducted a comprehensive subjective study involving 145 participants and six consumer-grade HDR-capable televisions. A total of over 22,000 pairwise comparisons were collected and scaled into Just-Objectionable-Difference (JOD) scores. Unlike prior datasets that focus on a single dynamic range format or use limited evaluation protocols, HDRSDR-VQA enables direct content-level comparison between HDR and SDR versions, supporting detailed investigations into when and why one format is preferred over the other. The open-sourced part of the dataset is publicly available to support further research in video quality assessment, content-adaptive streaming, and perceptual model development.
\end{abstract}


\section{Introduction}
High Dynamic Range (HDR) video has seen increasing adoption in consumer electronics and streaming services due to its ability to represent a broader range of luminance and color compared to Standard Dynamic Range (SDR). With support from modern display technologies and industry standards such as HDR10 and Dolby Vision, HDR promises more vivid and immersive visual experiences. In practice, the quality advantage of HDR content depends on a variety of factors, including content characteristics, encoding parameters, and most importantly, the display capabilities of the end device. For example, HDR content displayed on a low-end HDR-compatible device may suffer from suboptimal tone mapping or brightness limitations, potentially resulting in visual artifacts or unnatural appearance. Conversely, well-mastered SDR content can, in certain conditions, appear more stable and visually pleasing, especially in low-luminance or low-colorfulness scenes.

While a number of video quality assessment (VQA) datasets have been developed to support research on both SDR and HDR content, these resources tend to focus on one format and rarely provide paired comparisons between HDR and SDR versions of the same content. Moreover, existing studies often rely on rating methods such as Absolute Category Rating (ACR), which are not well suited to capturing subtle perceptual differences between similar video versions.

To address these limitations, we present HDRSDR-VQA, a large-scale video quality assessment dataset designed to enable fine-grained comparative analysis between HDR and SDR formats under real-world viewing conditions. The dataset comprises 960 videos derived from 54 diverse source sequences, each rendered in both HDR and SDR formats and further subjected to eight levels of distortion. Based on that, we conducted an extensive subjective study involving 145 participants and six different HDR-capable consumer televisions. Using pairwise comparison and active sampling, we collected over 22,000 comparisons and scaled them into Just-Objectionable-Difference (JOD) scores.

Through this work, we aim to provide the research community with a comprehensive resource for studying HDR and SDR video quality, benchmarking objective metrics, and developing new models for perceptual optimization in video encoding and delivery systems.

\section{Related Work}

Several subjective video quality databases have been developed over the past two decades, primarily for Standard Dynamic Range (SDR) content. Datasets such as~\cite{seshadrinathan2010study, moorthy2012video} include videos with various types of distortions and have been widely used for training and benchmarking VQA models. Most of these datasets, however, are limited to SDR and do not capture the visual properties unique to High Dynamic Range (HDR) content, such as high luminance, wide color gamut, and tone mapping artifacts.

Recent efforts have introduced HDR-specific VQA datasets~\cite{azimi2018evaluating, pan2018hdr, baroncini2016verification, rerabek2015subjective, athar2019perceptual}. LIVE-HDR~\cite{shang2023study} contains 310 videos derived from 31 reference sequences, covering multiple resolution and compression settings. LIVE-HDRvsSDR~\cite{hdrorsdr} further investigates perceptual preferences by including both HDR and SDR versions of the same content, using Absolute Category Rating (ACR) to collect user perception scores. While these datasets offer important insights, they have notable limitations. Most use fixed display conditions and a limited number of devices. Furthermore, the ACR method is not ideal for capturing fine-grained perceptual differences between HDR and SDR formats.

To our knowledge, no prior dataset provides large-scale pairwise comparison data across diverse HDR-capable TVs to directly quantify format preference between HDR and SDR under realistic conditions. Our work addresses this gap by offering a high-coverage database with 960 HDR/SDR videos, subjective scores collected from six different TVs, and detailed analysis on when and why HDR is preferred.

\begin{figure*}[htbp]
\begin{minipage}{\textwidth}
	\centering
	\subfigure{
		\begin{minipage}[b]{0.23\textwidth}
			\includegraphics[width=1\textwidth]{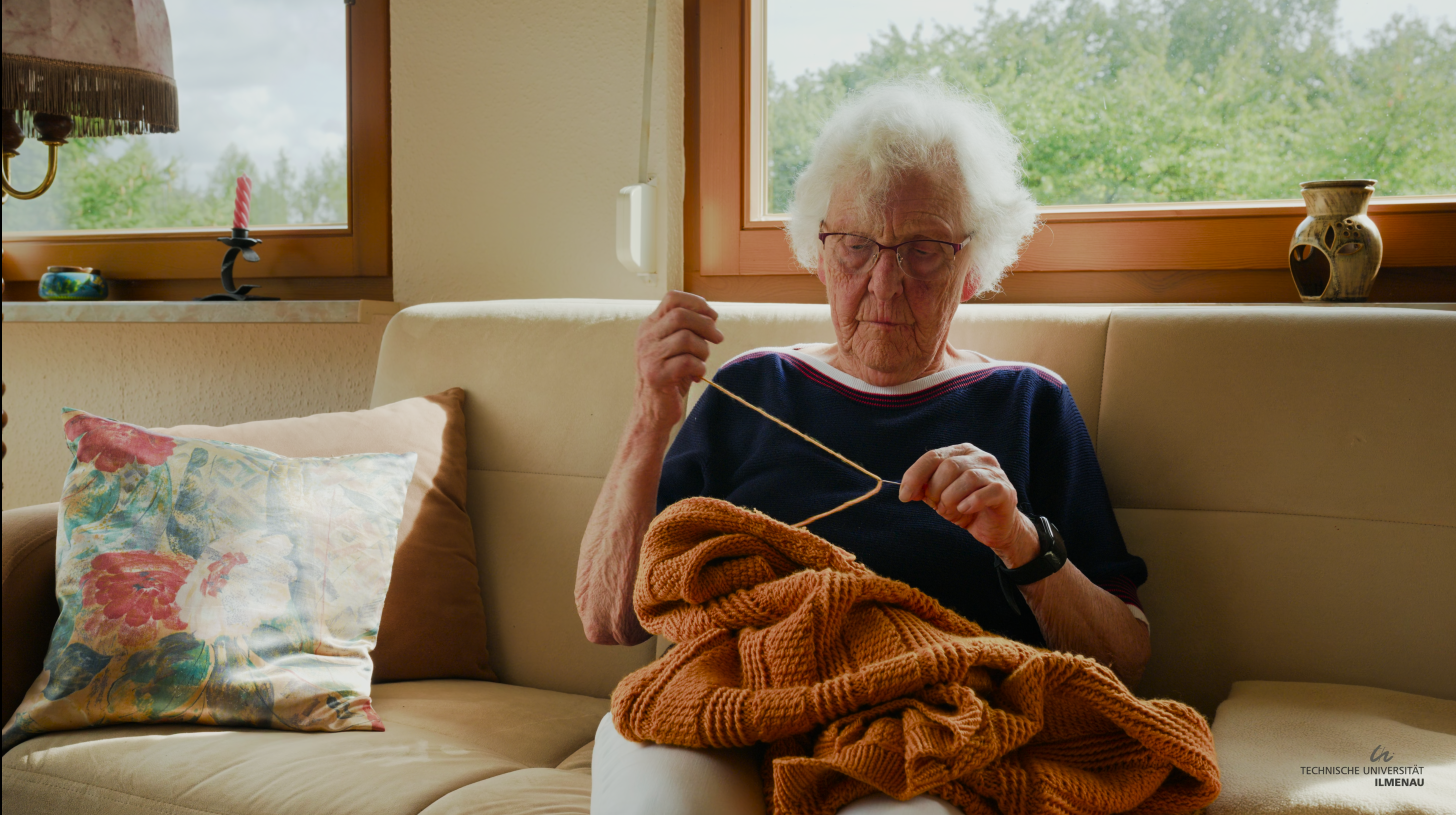}
		\end{minipage}
		\label{fig:hor_2figs_1cap_2subcap_1}
	}\hspace{-3mm}
    \subfigure{
    	\begin{minipage}[b]{0.23\textwidth}
   			\includegraphics[width=1\textwidth]{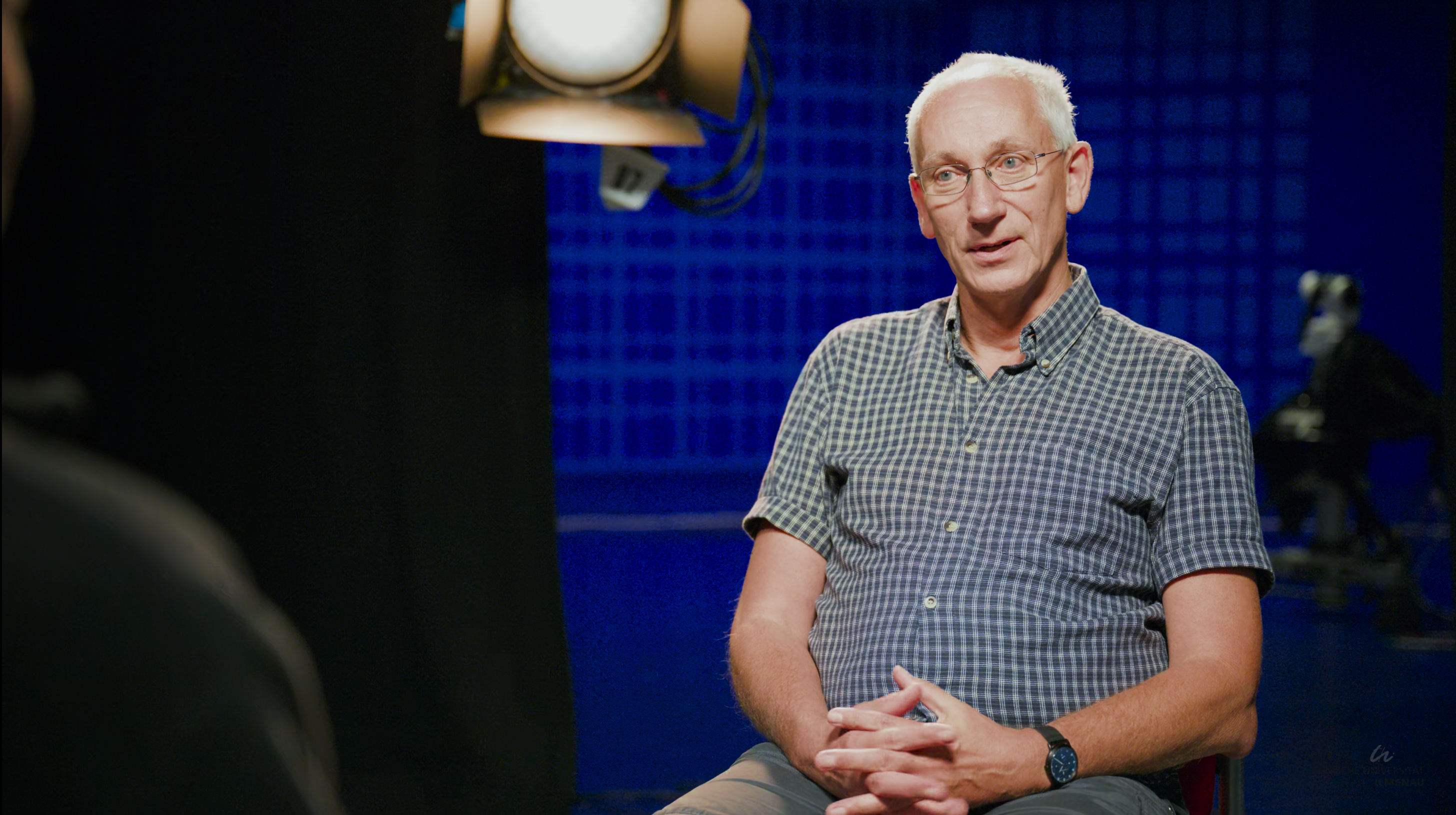}
    	\end{minipage}
		\label{fig:hor_2figs_1cap_2subcap_2}
    }\hspace{-3mm}
    \subfigure{
    	\begin{minipage}[b]{0.23\textwidth}
   			\includegraphics[width=1\textwidth]{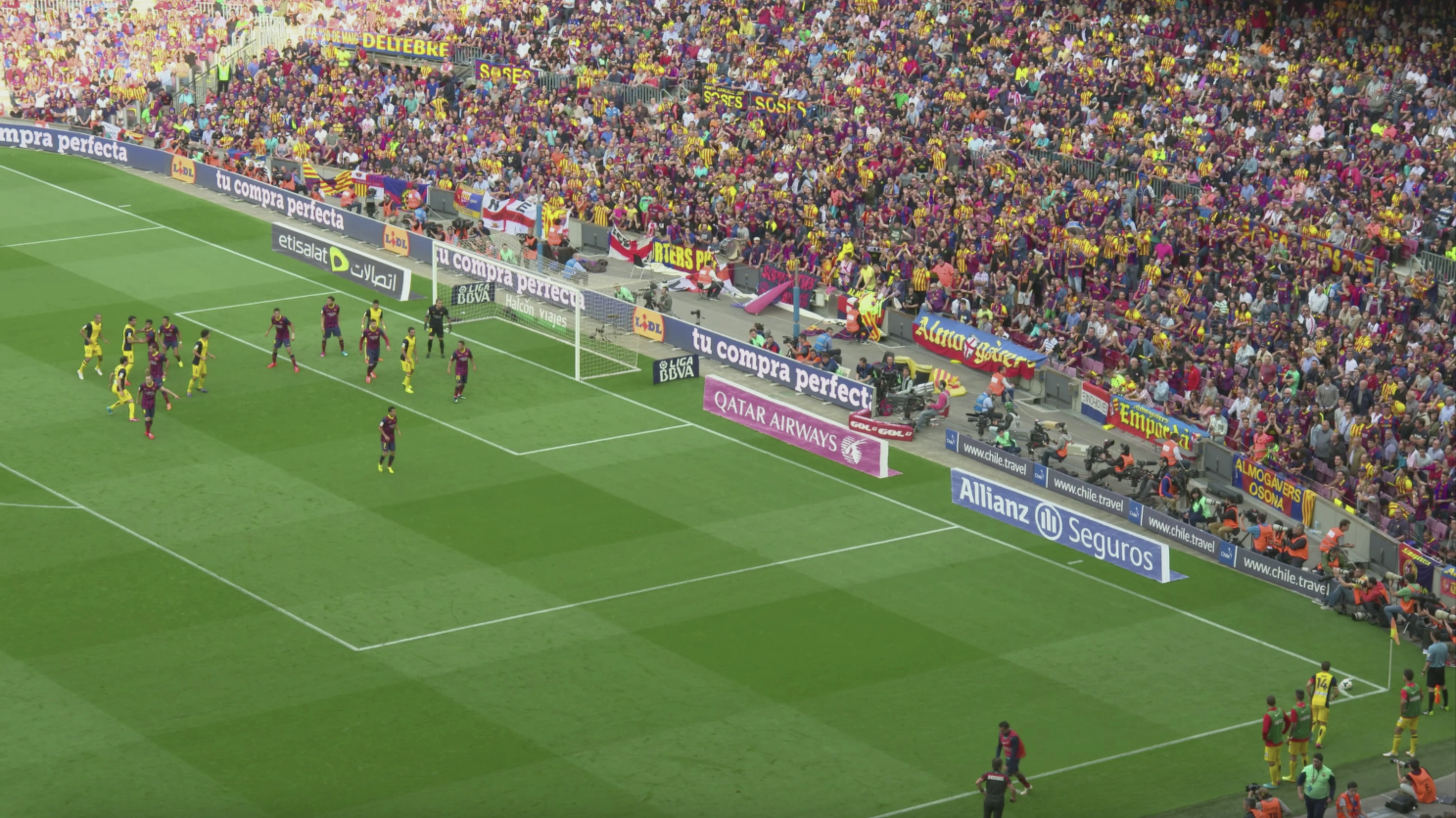}
    	\end{minipage}
		\label{fig:hor_2figs_1cap_2subcap_3}
    }\hspace{-3mm}
    \subfigure{
    	\begin{minipage}[b]{0.23\textwidth}
   			\includegraphics[width=1\textwidth]{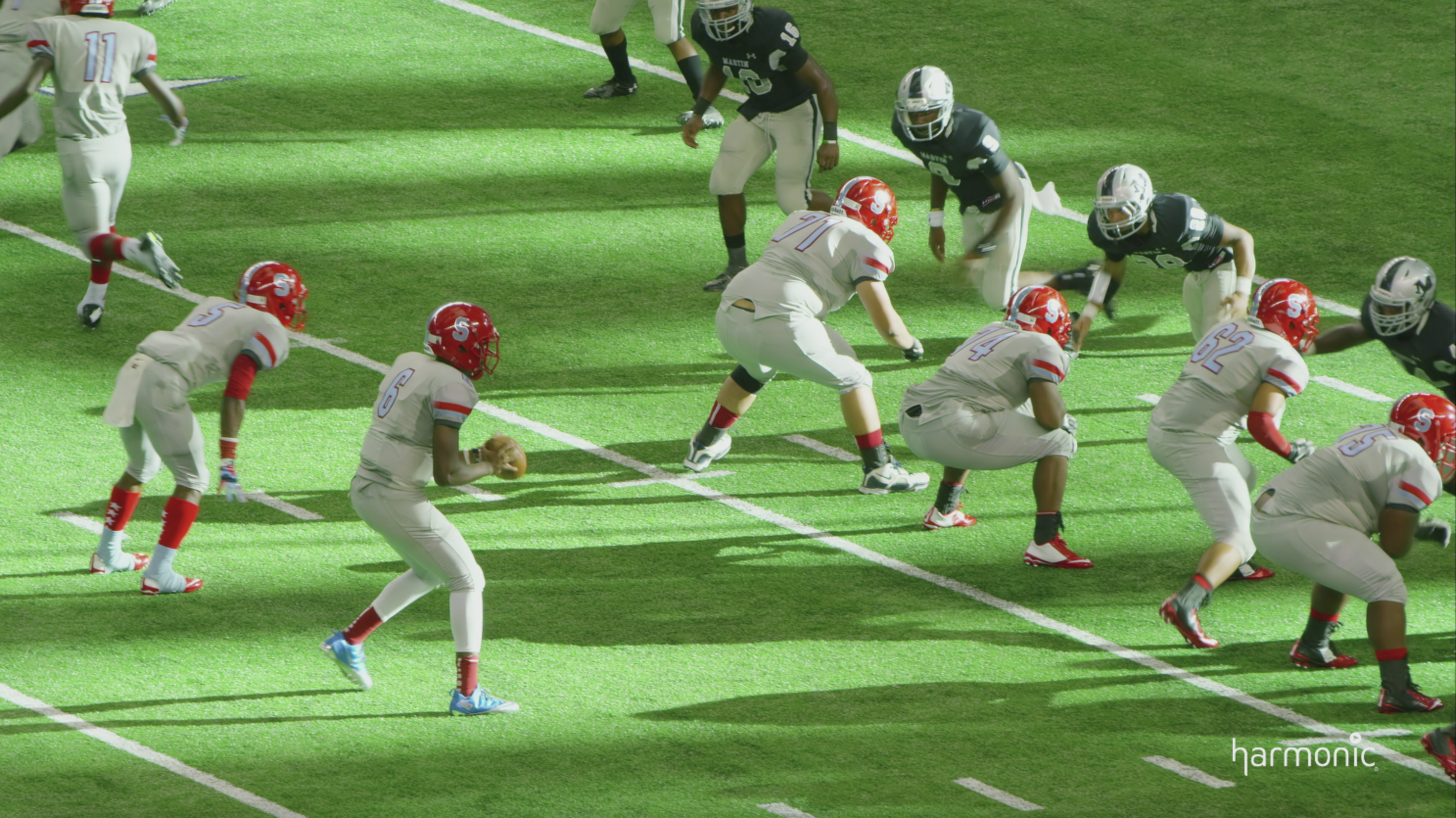}
    	\end{minipage}
		\label{fig:hor_2figs_1cap_2subcap_4}
    }

	\caption{Sample frames from the source sequences.}
	\label{fig1}
\end{minipage}
\end{figure*}
\footnotemark

\footnotetext{Due to copyright restrictions, actual frames from Amazon's live sports videos are not shown. We use screenshots from publicly available open-source sports videos as substitutes.}

\section{Database Construction}
\subsection{Source Sequences}

Figure \ref{fig1} presents sample frames from the source sequences in our database, showcasing a diverse range of video content. Our database contains 54 pristine high-quality or distortion-free videos. These videos include 31 open-source videos from the 8K HDR AVT-VQDB-UHD-2-HDR dataset \cite{AVT-VQDB-UHD-2-HDR}, 10 Video on Demand (VoD) videos and 10 Live Sports videos from Amazon Prime Video's internal source, and 3 anchor videos from the LIVE HDRorSDR database \cite{hdrorsdr}. All videos are in the BT.2020 \cite{bt2020} color gamut and are quantized using the PQ \cite{pq} Optical-Electronic Transfer Function (OETF). Each video sequence has a duration of approximately 7 seconds and includes static metadata of HDR10 standard. Detailed descriptions of the video categories are as follows:
\begin{itemize}
    \item \textbf{Open-Sourced Content:} This category comprises 31 pristine videos sourcd from the AVT-VQDB-UHD-2-HDR dataset \cite{AVT-VQDB-UHD-2-HDR}. The capture settings included 8.3K resolution, 12-bit color depth, 59.94 fps, and the NRAW codec (NEV format). The video file type was configured with a BT.2020 \cite{bt2020} color gamut and a PQ transfer function \cite{pq} for HDR. For this study, we encoded the pristine videos to 4K HDR10 format for use as source content.
    \item \textbf{VoD Content:} This category includes 10 professionally captured and graded videos designed for Video on Demand (VoD) streaming services. 
    \item \textbf{Live Sports Content:} The 10 Live Sports videos were captured by professional broadcasters during live events, including football, soccer and tennis games at stadiums. These videos reflect the high standards required for live sports broadcasting.
    \item \textbf{Anchor Contents:} Three contents from the LIVE HDRvsSDR database \cite{hdrorsdr} are included to calibrate and integrate data from this database with our current data set, further enhancing content quality and diversity.
\end{itemize}

\subsection{Content Analysis}

\begin{figure}[htbp]
\centering
\includegraphics[width=0.23\textwidth]{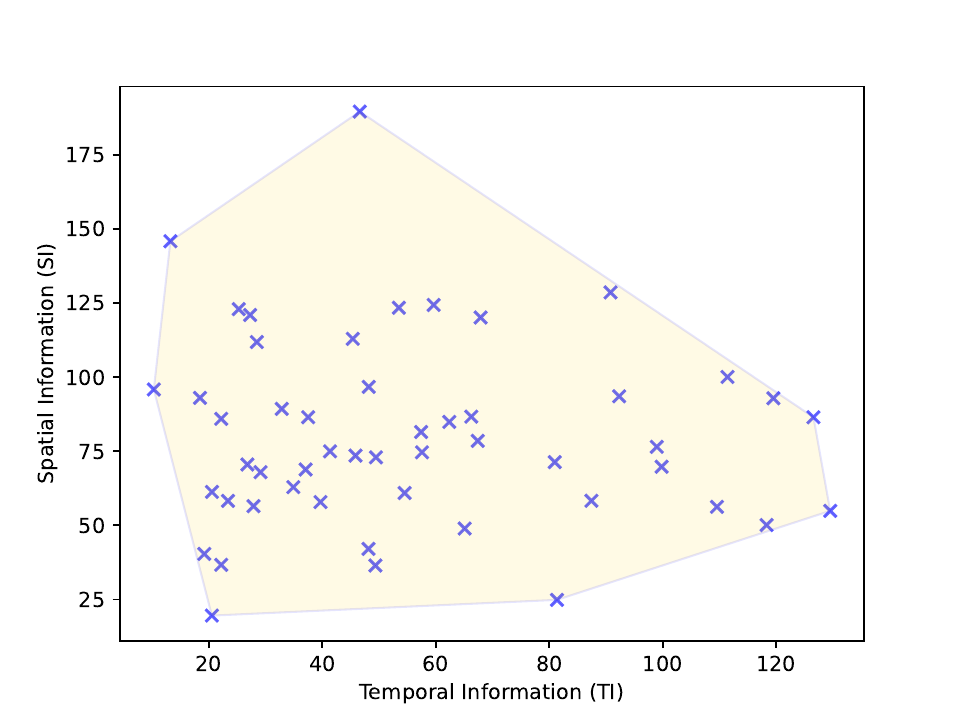}
\includegraphics[width=0.23\textwidth]{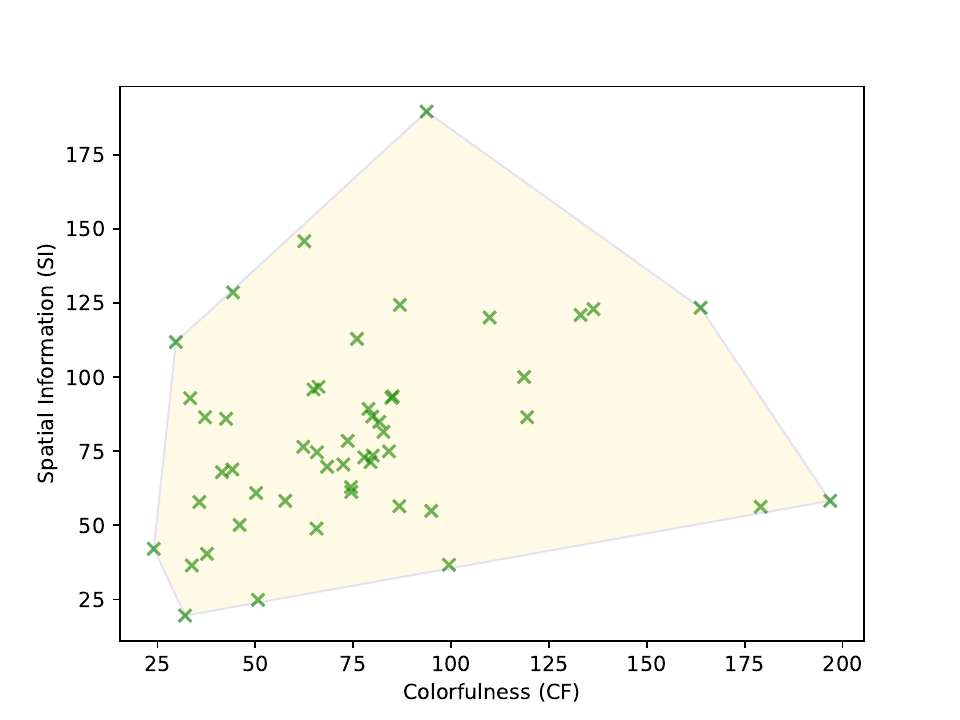}
\caption{Spatial Information (SI) plotted against Temporal Information (TI) and Spatial Information (SI) plotted against Colorfulness (CF) of the source videos except the anchor contents.}
\label{si_ti_cf}
\end{figure}

\begin{figure}[htbp]
\centering
\includegraphics[width=0.45\textwidth]{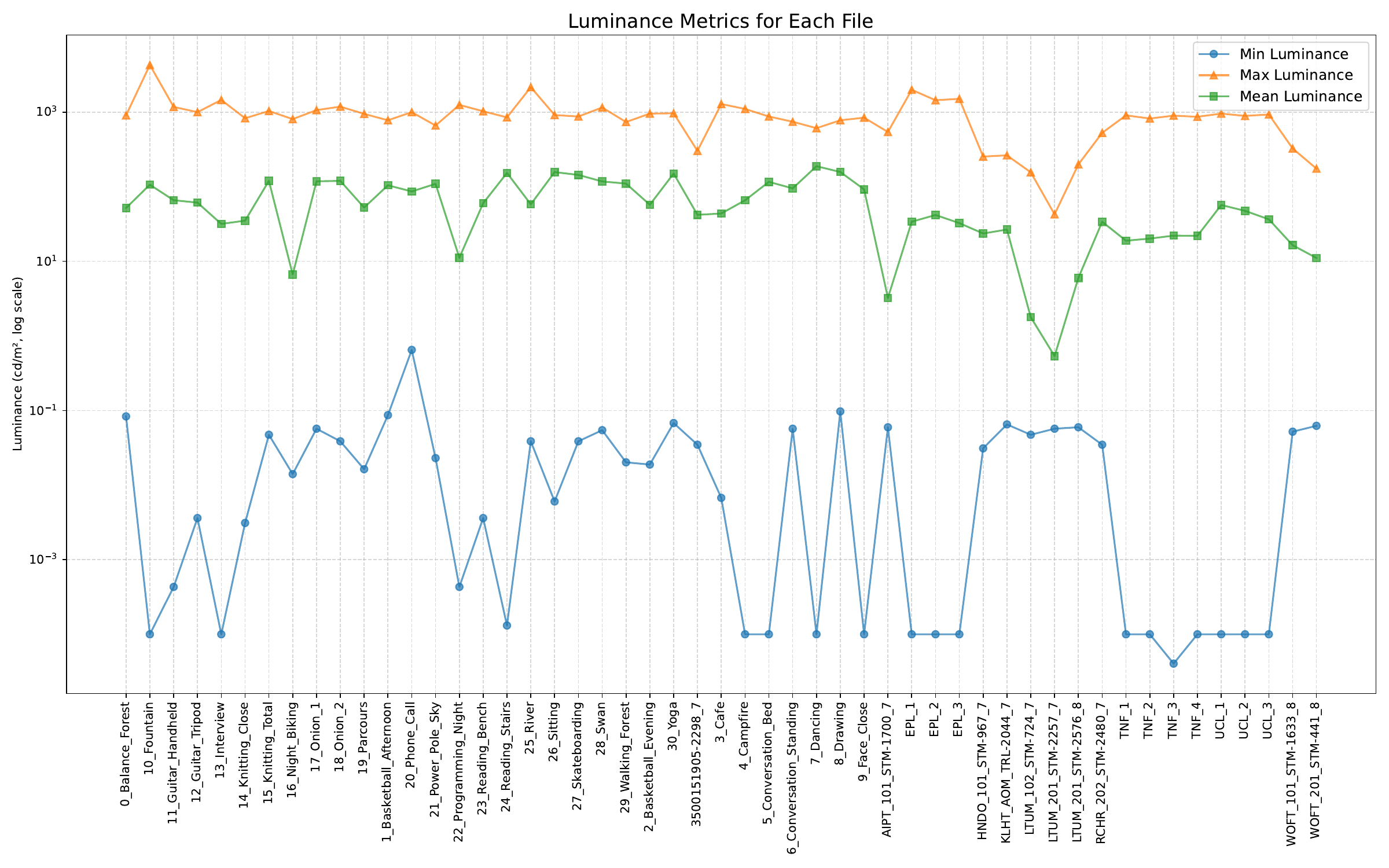}
\caption{Min, max, and mean luminance metrics measured on all of the source sequences except the anchor contents in the database.}
\label{luminance metrics}
\end{figure}

Figure \ref{si_ti_cf} illustrates the spatial information (SI), temporal information (TI), and colorfulness (CF) values for all source sequences, demonstrating a broad representation of spatial and temporal characteristics, as well as variations in colorfulness, across the dataset, while Figure \ref{luminance metrics} shows luminance metrics for each video in the dataset, including the minimum, maximum, and mean luminance values for each file. This visualization highlights the range of brightness levels across different videos, showcasing the diversity in luminance characteristics, which is crucial for understanding the content's visual attributes in both dark and bright scenes.

\subsection{Test Sequences}
All HDR10 source contents, except for the VoD videos, were converted to SDR format using the publicly available NBCU Lookup Tables (LUT). For VoD videos, both HDR and SDR versions were expertly created by Amazon Studios.

\begin{table}[htbp]
\caption{The distortion levels of the HDR and SDR video formats.\label{Distortion}}
\centering
\begin{tabular}{|c|c|c|c|}
\hline
Number & Resolution & Bitrate (kbps) \\ \hline
1 & 3840x2160 & 15000 \\ \hline
2 & 3840x2160 & 6000 \\ \hline
3 & 2560x1440 & 10000 \\ \hline
4 & 1920x1080 & 3000 \\ \hline
5 & 1280x720 & 4000 \\ \hline
6 & 1280x720 & 2000 \\ \hline
7 & 960x540 & 1500 \\ \hline
8 & 960x540 & 800 \\ \hline
\end{tabular}
\end{table}

In this study, we aim to comprehensively evaluate user perceptual differences caused by the most common viewing distortions, which arise from variations in bitrate and resolution levels. We introduced eight distinct levels of distortion applied to both HDR and SDR formats, in addition to one high-quality reference video. These distortion levels are carefully selected to ensure perceptual differentiation while ensure noticeable perceptual variation. The specific categories of distortions are detailed in Table \ref{Distortion}.

For encoding, we utilized the libx265 encoder in FFmpeg, operating in constant bitrate mode with single-pass encoding. This configuration is widely used in industry for its efficiency and straightforward implementation in streaming workflows. Each video sequence comprises 18 variations (9 HDR10 and 9 SDR), while the 3 anchor contents consist of 14 variations each, following the structure of the LIVE HDRorSDR database\cite{hdrorsdr}.

Together with the anchor contents, the database contains a total of 960 videos, equally divided between HDR10 (480 videos) and SDR (480 videos).

\section{Subjective Study Description}
\subsection{Display Devices}
Our experiments were conducted using 6 different 65" HDR10-compatible TVs to ensure a diverse evaluation across a range of display technologies. These TVs include the Samsung QN90B QLED (TV1), Samsung S95C OLED (TV2), Samsung CU8000 (TV3), TCL QM8/QM851G QLED (TV4), TCL Q7/Q750G QLED (TV5), and Vizio M6 Series Quantum 2022 (TV6). Below is an overview of each TV and its key features, followed by a discussion of the broader goal of achieving results representative of a general HDR TV. A detailed comparison of the TVs, including metrics such as peak brightness, BT.2020 color gamut coverage, and other key specifications, is presented in Table \ref{TVSpecs}, providing a comprehensive overview of their HDR performance characteristics.

\begin{table}[htbp]
\caption{A detailed comparison of the TVs, including metrics such as peak brightness, BT.2020 color gamut coverage. We refer to the peak luminance as the instantaneous brightness of a white rectangle displayed on an area covering 2\% of the screen. Rec 2020 Coverage ITP is the percentage of colors the TV can display, compared to the number of colors possible within the Rec.2020 gamut with a luminance range from 0 to 10,000 cd/m². \label{TVSpecs}}
\centering
\begin{tabular}{|c|c|c|}
\hline
TV model & \makecell{HDR Peak \\Brightness} &  \makecell{Rec.2020 \\Coverage ITP} \\ \hline
Samsung QN90B QLED & 1968 & 52.1\% \\ \hline
Samsung S95C OLED & 1229 & 53.7\% \\ \hline
Samsung CU8000 & 295 & 23.8\% \\ \hline
TCL QM8/QM851G QLED & 2564 & 54.8\% \\ \hline
TCL Q7/Q750G QLED & 381 & 28.7\% \\ \hline
Vizio M6 Series Quantum 2022 & 252 & 26.1\% \\ \hline
\end{tabular}
\end{table}

Our selection of TVs covers a wide range of display technologies, price points, and performance levels, from high-end QLED and OLED models to more budget-friendly options. By conducting experiments across this diverse set of devices, we aim to capture a comprehensive understanding of how HDR content is perceived on general types of TVs.

The ultimate goal is to ensure that the results of our study are not limited to specific devices but are generalizable to the broader category of HDR-compatible TVs. This is crucial as HDR content is consumed on a wide variety of displays, each with its unique capabilities and limitations. The diverse testing setup allows us to derive insights that are representative of the general HDR viewing experience, ensuring the applicability of our findings to a wider audience and broader range of devices.

Each TV is connected to an Amazon Fire TV stick. We employed an Android interface equipped with the EXO Player. This method ensures a standardized environment for data collection across all configurations.

\subsection{Subjects}
A total of 145 participants were recruited from the public to complete the study at the LIVE Lab at the University of Texas at Austin. The participant pool included 53 females and 91 males, with the following age distribution: 5 participants were under 20, 112 were aged 20-30, 21 were aged 30-40, and 5 were over 40. One participant chose not to disclose their gender and age. Additionally, one subject was identified as color deficient; however, they were not excluded from the study in line with our practice of accommodating diverse participants, which follows the crowd-sourcing approach. All participants were required to wear glasses during the study if they normally use them, ensuring consistent visual conditions and accurate representation of their usual viewing experiences.

Ultimately, 19 participants completed the study on TV1, 26 on TV2, 21 on TV3, 26 on TV4, 26 on TV5, and 27 on TV6, ensuring comprehensive data collection across all tested devices.

\subsection{Ambient Conditions}
Our strategy involves adopting a home viewing environment, in alignment with one of the best practice recommendations for critical evaluation of HDR content. In this environment, a set of studio LED lights will be set to produce an incident illumination on the TV around 200 lux. This setting is designed to optimize the visual experience and ensure the most accurate assessment of HDR performance.

\subsection{Study Design}
Our study design employed a pairwise comparison (PC) approach, following the guidelines outlined in ITU-R BT 500.13 \cite{bt500}, in combination with the Active Sampling for Pairwise Comparisons (ASAP) algorithm \cite{asap}, to efficiently and accurately assess subjective video quality.

\begin{figure}[htbp]
\centering
\includegraphics[width=0.45\textwidth]{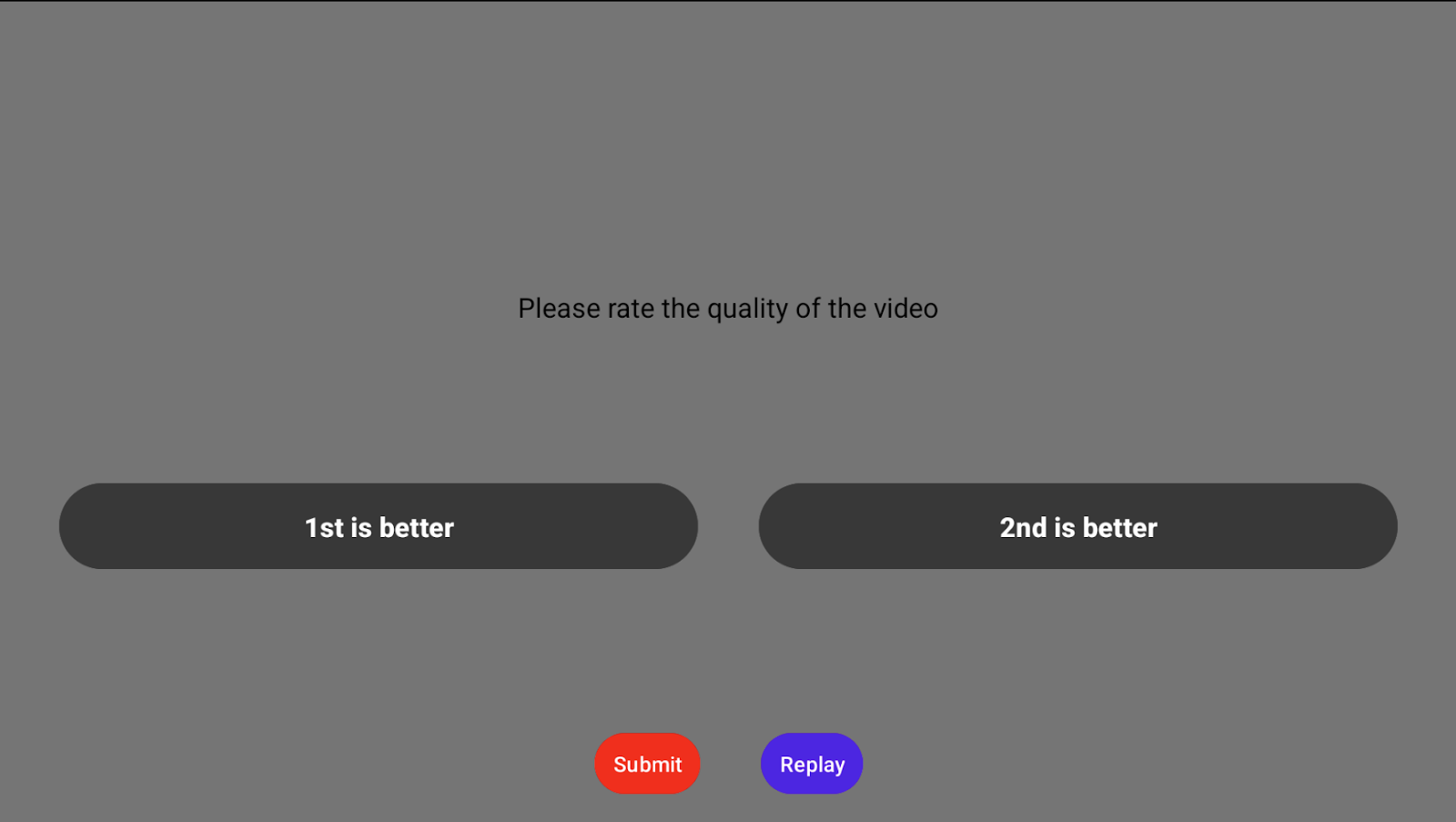}
\caption{Screenshot of the rating screen used to determine which video sequence has higher perceived quality.}
\label{rating}
\end{figure}

In the PC method, participants are presented with two video sequences of the same content and are asked to select the one with higher perceived quality, with the rating interface illustrated in Figure \ref{rating}. Participants can replay the videos to ensure an informed decision. This approach is widely recognized for its simplicity and reliability, as it avoids the complexities of traditional rating scales and reduces variability in judgments across participants.

To further improve efficiency, we incorporated the ASAP algorithm, an active sampling framework that dynamically identifies the most informative video pairs by analyzing uncertainty in the current ranking model. By prioritizing comparisons with the greatest potential to refine quality estimates, ASAP reduces redundancy, accelerates convergence to accurate rankings, and minimizes the number of comparisons required. During the study, participants viewed video pairs selected by ASAP. 

Before the study began, participants underwent a training session designed to familiarize them with the experimental setup and rating process. The training session involved videos that were not part of the study database, including pairs with varying levels of compression to represent the quality range in the dataset. Participants were shown how to use the TV controller to make their selections, and clear instructions were given to base their ratings solely on video quality. No additional details about the study's purpose were disclosed to avoid introducing bias.

Each participant evaluated 160 video pairs in a single session, which lasted approximately one hour. To ensure comfort and maintain focus, participants were encouraged to take breaks whenever they felt tired during the session. 

We utilized the \textit{pwcmp} algorithm \cite{pwcmp} to perform pairwise scaling, transforming the raw pairwise comparison data into JODs for each video. The difference of 1 JOD indicates that 75\% of observers selected one condition as better than the other. The \textit{pwcmp} algorithm employs a maximum likelihood estimation framework to infer a continuous quality scale based on the comparison results. This method accounts for the probabilistic nature of human judgments, incorporating inconsistencies and uncertainties in the pairwise data. 

\section{Subjective Score Analysis}

We observed that there are significant differences in the subjective quality scores of different video content in HDR and SDR formats. Specifically, in some high-brightness and high-saturation videos, HDR shows a stable and significant subjective advantage; while in scenes with limited brightness or color distribution and high motion intensity, SDR sometimes performs better.

To further understand the impact of content attributes on subjective preferences, we selected four representative video contents, representing typical scenes such as high texture/high brightness, high dynamic, low brightness/low color, and high motion complexity, and analyzed the differences in their HDR and SDR performance one by one.

\begin{figure}[tbp]
\centering
\subfigure[0\_Balance\_Forest]{
\includegraphics[width=0.22\textwidth]{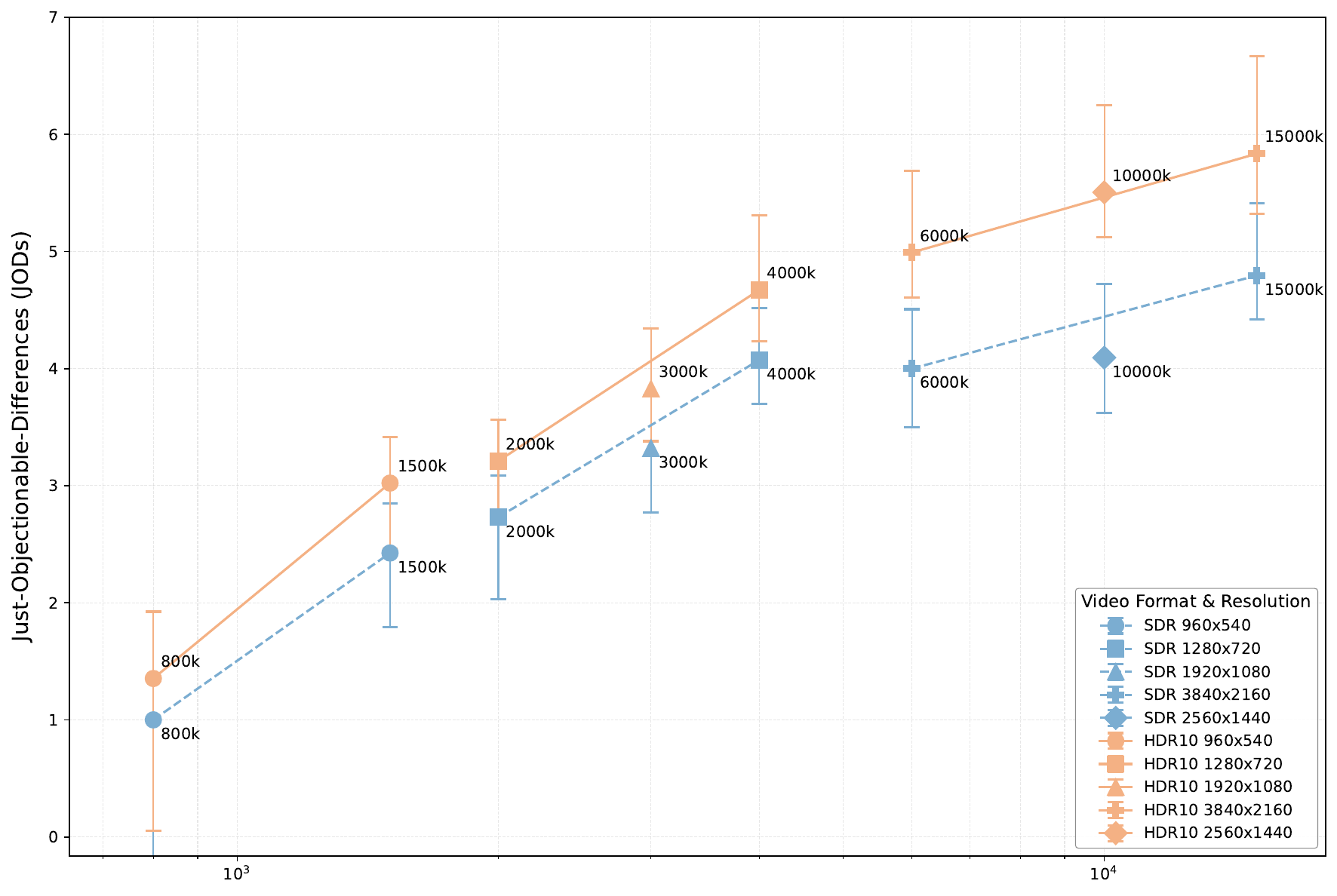}
}
\subfigure[28\_Swan]{
\includegraphics[width=0.22\textwidth]{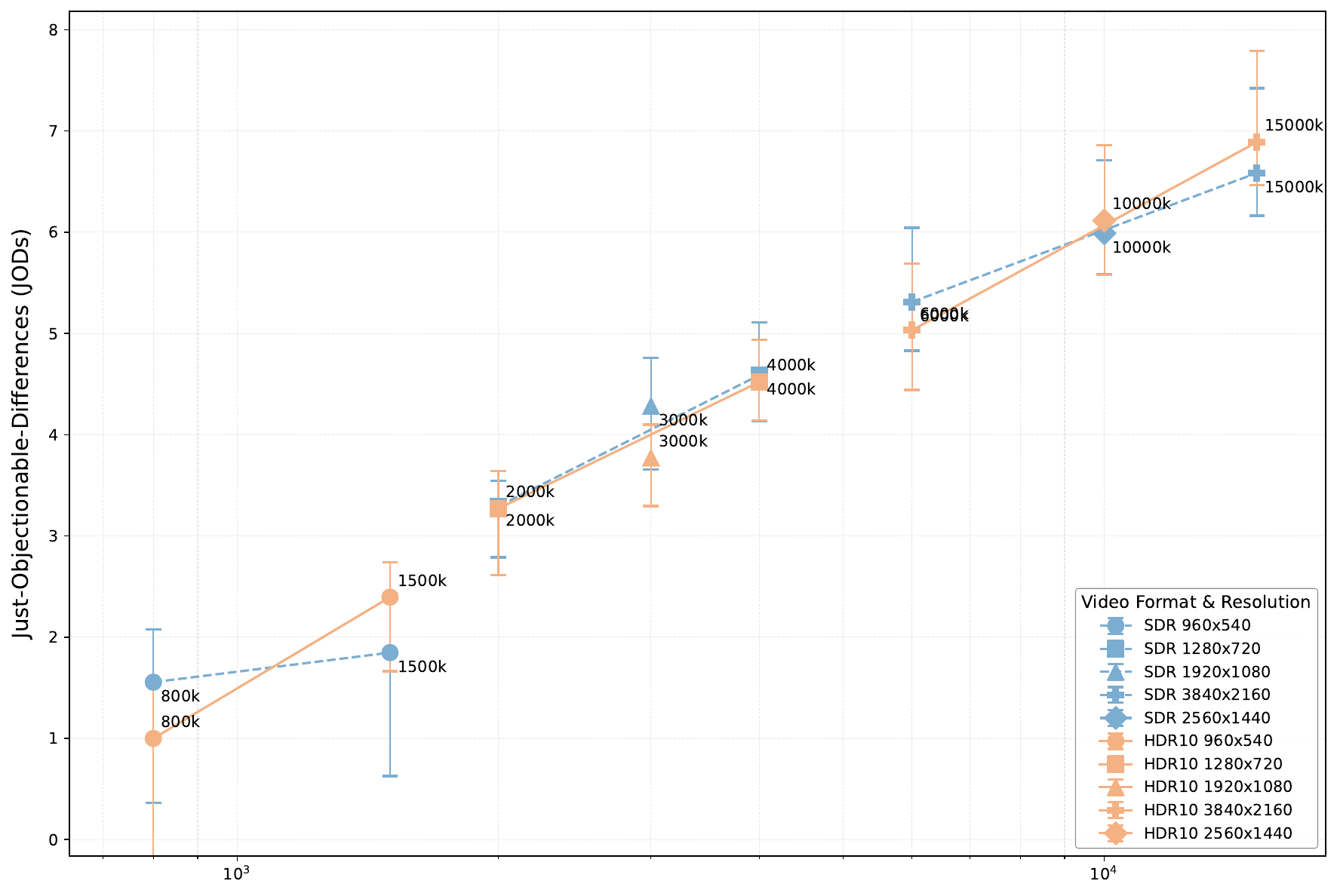}
}
\subfigure[WOFT\_201\_STM-441\_8]{
\includegraphics[width=0.22\textwidth]{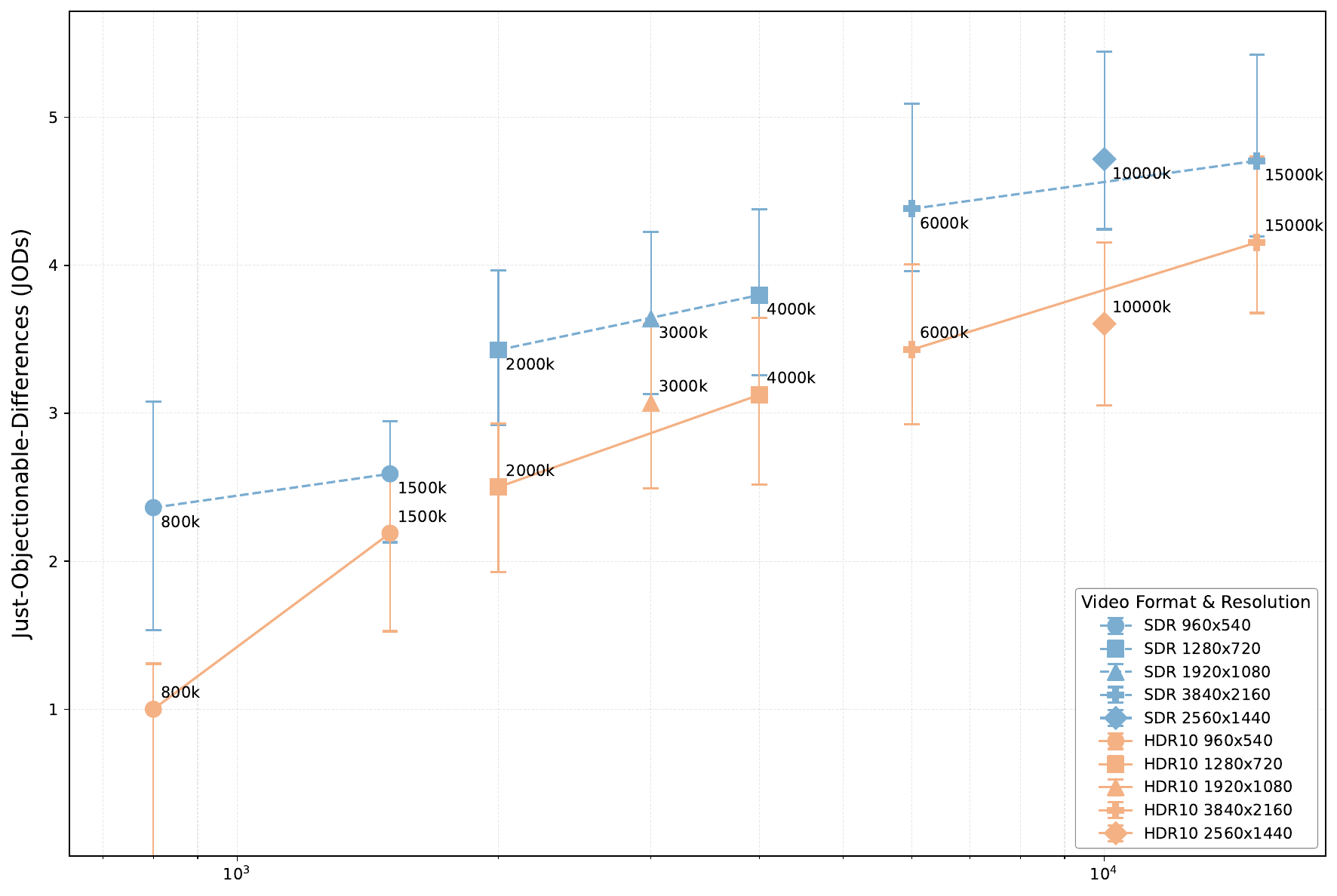}
}
\subfigure[EPL\_1]{
\includegraphics[width=0.22\textwidth]{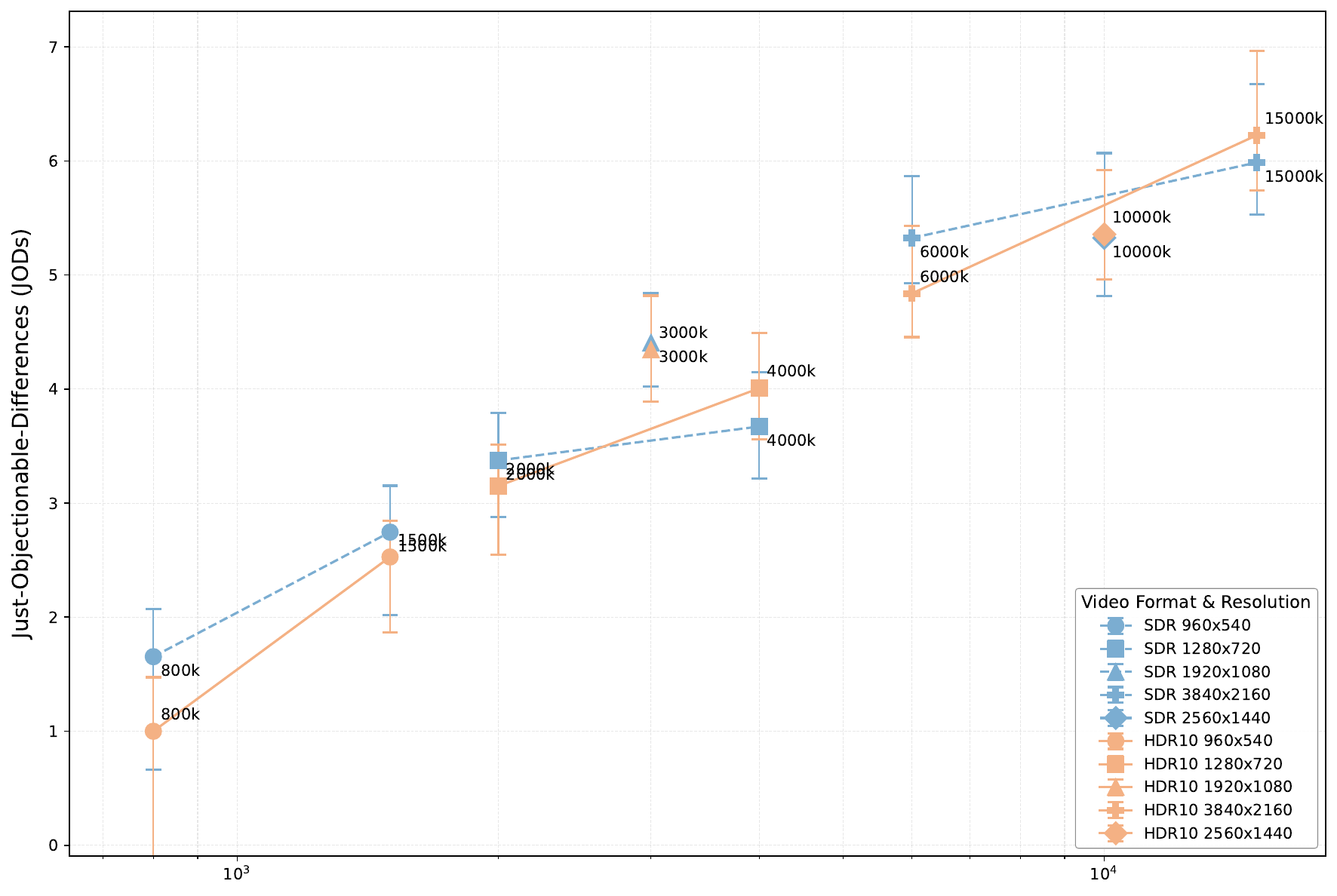}
}
\caption{Examples Rate-Distortion Curve of four contents collected in our subjective study. }
\label{fig:HDR-SDR screenshots}
\end{figure}

"0\_Balance\_Forest" is an open-source content containing rich texture details. It has high spatial information (SI = 122.96), indicating significant visual complexity. The dynamic information is moderate (TI = 25.28) with gentle scene transitions. Additionally, 17.29\% of the pixels exceed the SDR brightness range, showcasing distinct highlight areas. Furthermore, 23.54\% of the pixels are outside the sRGB gamut, indicating that HDR can present more color details in a wide color gamut. HDR10 consistently outperforms SDR across all bitrates, with its performance advantage becoming more pronounced as the bitrate increases. The high bit depth of HDR10 ensures refined brightness rendering in highlight areas and significantly better detail preservation compared to SDR. Even at low bitrates, HDR10 leverages its advanced technical features to maintain high-quality performance, while at high bitrates, its advantages are even more significant.

"28\_Swan" is characterized by high temporal information (TI = 119.51), reflecting significant motion or rapid scene changes. It also has a high proportion of pixels (31.43\%) exceeding the SDR brightness range, indicating high brightness levels. Across all bitrates, the difference between HDR and SDR JOD is minimal. At low bitrates, SDR scores slightly higher than HDR, but the difference is negligible. As bitrate increases, the scores of HDR and SDR remain very close, with no significant difference observed even at 2000kbps to 6000kbps. At the highest bitrates, HDR scores are only slightly higher than SDR. The high temporal complexity likely masks the potential advantages of HDR, as frequent motion and changes dominate the perception of quality.

"WOFT\_201\_STM-441\_8" is one of the Amazon VoD content, which has an extremely low proportion of pixels exceeding the SDR brightness range (0.003\%), indicating an absence of significant highlight areas. The color gamut is similarly limited, with only 0.006\% of pixels outside the sRGB range, reflecting a lack of color diversity. At low bitrates, SDR scores slightly higher than HDR in JOD. The higher bit depth of HDR may contribute to less stable detail retention under limited bitrate conditions. The lack of significant brightness and color diversity further undermines HDR’s advantages at these bitrates. As the bitrate increases, the score gap narrows, although SDR maintains a slight edge. At the highest bitrates, SDR continues to score marginally higher, as the video’s inherent characteristics prevent HDR from leveraging its full technical potential.

"EPL\_1" features soccer game with high spatial information (SI = 124.38) and temporal information (TI = 59.62), indicating detailed textures, frequent motion, and scene changes. However, the proportions of pixels exceeding the sRGB color gamut (0.003\%) and SDR brightness range (0.016\%) are minimal. At low bitrates and resolutions, SDR scores slightly higher than HDR. This can be attributed to HDR's higher bit depth requirements, which may not be fully supported under low-bitrate compression. Additionally, the lack of significant highlight areas limits HDR’s brightness expansion advantage. As bitrate and resolution increase, HDR and SDR scores converge, with HDR slightly outperforming SDR at higher bitrates and resolutions. HDR demonstrates better detail restoration, particularly in grass textures and player dynamics, but the overall improvement remains modest.

\section{Conclusion}

In this work, we introduced HDRSDR-VQA, a large-scale subjective video quality assessment dataset designed to facilitate direct comparison between HDR and SDR content under realistic viewing conditions. The dataset includes 960 videos derived from 54 diverse source sequences, presented in both HDR10 and SDR formats across nine quality levels. Subjective evaluations were conducted using pairwise comparisons on six HDR-capable consumer TVs, resulting in over 22,000 judgments from 145 participants. Our analysis demonstrates that HDR generally offers perceptual advantages in content with rich textures and high brightness or color gamut, but these advantages diminish or even reverse in low dynamic or motion-intensive scenes. We expect this dataset to serve as a valuable resource for future research in video quality assessment, HDR encoding, and perceptual optimization for adaptive streaming systems.

\bibliography{paper}
\bibliographystyle{IEEEtran}

\end{document}